\documentclass{article}

\usepackage{mathptmx}
\usepackage{fullpage}
\usepackage{parskip}

\usepackage{appendix}
\usepackage{amsfonts}
\usepackage{amsmath}
\usepackage{amssymb}
\usepackage{amsthm}
\usepackage{mathtools}
\usepackage{appendix}
\usepackage{color}
\usepackage{float}
\usepackage{graphicx}
\usepackage{ltxtable}
\usepackage{natbib}
\usepackage{nicefrac}
\usepackage{subfigure}
\usepackage{wrapfig}
\usepackage{url}
\usepackage{hyperref}

\newcolumntype{L}{>{\centering\arraybackslash} m{0.04\columnwidth}}
\newcolumntype{H}{>{\centering\arraybackslash} m{0.49\textwidth}}
\newcolumntype{S}{>{\centering\arraybackslash} m{0.28\textwidth}}
\newcolumntype{T}{>{\centering\arraybackslash} m{0.31\textwidth}}

\newcommand{\todo}[1]{ \textcolor{red}{TODO: #1} }

\newcommand{\inner}[1]{\langle {#1} \rangle}

\newcommand{\E}{\ensuremath{\mathbb{E}}}

\newcommand{\R}{{\mathbb{R}}}

\newtheorem{theorem}{Theorem}
\newtheorem{claim}{Claim}

\newcommand{\X}{\mathcal{X}}
\newcommand{\Y}{\mathcal{Y}}
\newcommand{\x}{\mathbf{x}}
\newcommand{\w}{\mathbf{w}}
\newcommand{\bphi}{\boldsymbol{\phi}}
\newcommand{\bphia}{\tilde{\bphi}}
\newcommand{\phia}{\tilde{\phi}}
\renewcommand{\H}{\mathcal{H}}

\newcommand{\bomega}{\boldsymbol{\omega}}

\title{Explicit Approximations of the Gaussian Kernel}

\author{
	Andrew Cotter, Joseph Keshet and Nathan Srebro\\
	\small\sc{\{cotter,jkeshet,nati\}@ttic.edu}\\
	\small\it{Toyota Technological Institute at Chicago}\\
	\small\it{6045 S. Kenwood Ave.}\\
	\small\it{Chicago, Illinois 60637, USA}
}

\begin{document} 

\maketitle

\begin{abstract}
We investigate training and using Gaussian kernel SVMs by approximating the kernel
with an explicit finite-dimensional polynomial feature representation based on
the Taylor expansion of the exponential. Although not as efficient as the
recently-proposed random Fourier features \citep{RahimiRe07} in terms of the
{\em number} of features, we show how this polynomial representation can
provide a better approximation in terms of the {\em computational cost}
involved. This makes our ``Taylor features'' especially attractive for use on
very large data sets, in conjunction with online or stochastic training.
\end{abstract}

\section{Introduction}\label{sec:introduction}

In recent years several extremely fast methods for training {\em linear}
support vector machines have been developed. These are generally stochastic
(online) methods, which work on one example at a time, and for which each step
involves only simple calculations on a single feature vector: inner products
and vector additions \citep{ShalevSiSr07,HsiehChLiKeSu08}. Such methods are
capable of training support vector machines (SVMs) with many millions of
examples in a few seconds on a conventional CPU, essentially eliminating any
concerns about training runtime even on very large datasets.

Meanwhile, fast methods for training kernelized SVMs have lagged behind.
State-of-the-art kernel SVM training methods may take days or even weeks of
conventional CPU time for problems with a million examples of effective
dimension less than 100. While the stochastic methods mentioned above can
indeed be kernelized, each iteration then requires the computation of an entire
row of the kernel matrix, i.e.~the entire data set needs to be considered in
each stochastic step.

Any Mercer kernel implements an inner-product between a mapping of two input
vectors into a high dimensional feature space. In this paper we propose an
explicit low-dimensional approximation to this mapping, which, after being
applied to the input data, can be used with an efficient linear SVM solver.
The dimension of the approximate mapping controls the computational difficulty
and the approximation qualities. The key to choosing a good approximate mapping
comes in trading off these considerations.

\citet{RahimiRe07} proposed such a feature representation for the Gaussian
kernel (as well as other shift-invariant kernels) using random ``Fourier''
features: each feature (each coordinate in the feature mapping) is a cosine of
a random affine projection of the data.

In this paper we study an alternative simple feature representation
approximating the Gaussian kernel: we take a low-order Taylor expansion of the
exponential, resulting in features that are scaled monomials in the coordinates
of the input vectors. We focus on the Gaussian kernel, but a similar approach
could also work for other kernels which depend on distances or inner products
between feature vectors, e.g. the sigmoid kernel.

At first glance it seems that this Taylor feature representation must be
inferior to random Fourier features. The theoretical guarantee on the
approximation quality is given by the error of a Taylor series, and is
expressed most naturally in terms of the degree of the expansion, of which the
number of features is an exponential function. Indeed, to achieve the same
approximation quality, we need many more Taylor than random Fourier features
(see Section \ref{sec:comparison} for a detailed analysis). Furthermore, the
Taylor features are {\em not} shift and rotation invariant, even though the
Gaussian kernel itself is of course shift and rotation invariant.

However, we argue that when choosing an explicit feature representation, one
should focus not on the {\em number} of features used by the representation,
but rather on the computational cost of computing it. In online (or stochastic)
optimization, each example is considered only once, or perhaps a few times, and
the cost to the SVM optimizer of each step is essentially just the cost of
reading the feature vector. Even if each training example is considered several
times, the dataset will often be sufficiently large that precomputing and
saving all feature vectors is infeasible. For example, consider a data set of
hundreds of millions of examples, and an explicit feature mapping with 100,000
features. Although it might be possible to store the input representation in
memory, it would require tens of terabytes to store the feature vectors.
Instead, one will need to re-compute each feature vectors when required. The
computational cost of training is then dominated by that of the computing the
feature , and we should judge the utility of a feature mapping not by the
approximation quality as a function of dimensionality, but rather as a function
of computational cost.

We will discuss how the cost of computing the Taylor features can be
dramatically less than that of the random Fourier features, especially for
sparse input data. In fact, the advantage of the Taylor features over the
random Fourier features for sparse data is directly related to the Taylor
features {\em not} being rotationally and shift invariant, as these operations
do not preserve sparsity. We demonstrate empirically that on many benchmark
datasets, although the Taylor representation requires many more features to
achieve the same approximation quality as random Fourier features, it
nevertheless outperforms a random Fourier features in terms of approximation
and prediction quality as a function of the computational cost.

{\bf Related Work} 
\citet{FineSc02} and \citet{BalcanBlVe06} suggest obtaining a low-dimensional
approximation to an arbitrary kernel by approximating the empirical Gram
matrix. Such approaches invariably involve calculating a factorization of (at
least a large subset of) the Gram matrix, an operation well beyond reach for
large data sets. Here, we use an efficient non-data-dependent approximation
that relies on analytic properties of the Gaussian kernel.

A similar approximation of the Gaussian kernel by a low-dimensional Taylor
expansion was proposed by \citet{YangDuDa04}, who used this approximation to
speed up a conjugate gradient optimizer. \citet{XuPoJePr06} also proposed the
use of the Taylor expansion to explicitly approximate the Hilbert space induced
by the Gaussian kernel, but presented neither experiments nor a quantitative
discussion of the approximation. We are not aware of any comparison of the
Fourier features with the Taylor features, beyond a passing mention by
\citet{RahimiRe08} that the number of Taylor features required for good
approximation grows rapidly. In particular, we are not aware of a previous
analysis taking into account the computational cost of generating the features,
which is an important issue that, as we discuss here, changes the picture
entirely.

\section{Kernel Projections and Approximations}\label{sec:preliminaries}

Consider a classifier based on a predictor $f:\X\to\R$, which is trained by
minimizing the regularized training error on a training set of examples
$S=\{\x_i,y_i\}_{i=1}^m$, where $\x_i\in \X$ and $y_i\in\Y$. Here we take $\Y =
\{\pm 1\}$, and minimize the hinge loss, although our approach holds for other
loss functions, including multiclass and structured loss.

The ``kernel trick'' is a popular strategy which permits using linear
predictors in some implicit Hilbert space $\H$, i.e. predictors of the form
$f(\x)=\inner{\w,\bphi(\x)}$, where $\|w\|_{\H}$ is regularized, and
$\bphi:\X\to\H$ is given implicitly in terms of a {\em kernel function}
$K(\x,\x')=\inner{\bphi(\x),\bphi(\x')}$. The Representer Theorem guarantees
that the predictor minimizing the regularized training error is of the form:
\begin{equation}\label{eq:func1}
f^{*}(\x) = \sum_{i=1}^m \alpha_i \inner{\bphi(\x_i),\bphi(\x)} = \sum_{i=1}^m
\alpha_i K(\x_i,\x)
\end{equation}
for some set of coefficients $\alpha_i\in\R$. It suffices then, to search over
the coefficients $\alpha_i\in\R$ when training. However, when the size of
the training set, $m$, is very large, it can be very expensive to evaluate
\eqref{eq:func1} for even a single $\x$. For example, for a $d$ dimensional
input space $\X=\R^d$, and with a kernel whose evaluation runtime is even
just linear in $d$ (e.g.~the Gaussian kernel, as well as most other simple
kernels), evaluating \eqref{eq:func1} requires $O(d \cdot m)$ operations.
The goal of this paper is to study an explicit finite dimensional
approximation $\bphia:\R^d\to\R^D$ to the mapping $\bphi$, which alleviates
the need to use the representation \eqref{eq:func1}. We will then consider
classifiers $\tilde{f}$ of the form:
\begin{equation*}
\tilde{f}(\x) = \inner{\tilde{\w},\bphia(\x)}
\end{equation*}
where $\tilde{w}\in\R^D$ is a weight vector which we represent explicitly.
Evaluating $\tilde{f}(\x)$ requires $O(D)$ operations, which is better than the
representation \eqref{eq:func1} when $D \ll d\cdot m$.

One option for constructing such an approximation is to project the
mapping $\phi$ onto a $D$-dimensional subspace of $\H$:
$\bphia(\x)=P\bphi(\x)$. This raises the question of how one may most
effectively reduce the dimensionality of the subspace within which we
work, while minimizing the resulting approximation error. Our first
result will bound the error which results from solving the SVM problem
on a subspace of $\H$.

Consider the kernel Support Vector Machines (SVM) optimization problem (using $(\cdot)_{+}$ to denote $\max\{0,\cdot\}$):
\begin{equation}\label{eq:svm}
\min_{\w\in\H} p(\w) ~=~ \frac{\lambda}{2} \|\w\|_{\H}^{2} + \frac{1}{m}
\sum_{i=1}^{m} \left( 1-y_i \inner{\w,\bphi(\x_i)}\right)_{+}
\end{equation}
and denote by $\tilde{p}(\tilde{\w})$ the objective function which results from
replacing the mapping $\bphi$ with the approximate mapping $\bphia$. Recall
also that $K(\x,\x')=\inner{\bphi(\x),\bphi(\x)}$ and denote
$\tilde{K}(\x,\x')=\inner{\bphia(\x),\bphia(\x')}$

\begin{theorem}\label{thm:primal_objective}
Let $p^{*}=\inf_{\w} p(\w)$ be the optimum value of \eqref{eq:svm}. For any
approximate mapping $\bphia(\x)=P\bphi(x)$ defined by a projection $P$, let
$\tilde{p}^{*}=\inf_{\tilde{\w}} \tilde{p}(\tilde{\w})$ be the optimum value of
the SVM with respect to this feature mapping. Then:
\begin{equation*}
p^{*} \le \tilde{p}^{*} \le p^{*} + \frac{1}{m\sqrt{\lambda}} \sum_{i=1}^{m}
\sqrt{ K( \x_i, \x_i ) - \tilde{K} ( \x_i, \x_i ) }
\end{equation*}
\end{theorem}
Note that since we also have $\|\bphia(\x)\| \leq \|\bphi(\x)\|$, it is
meaningful to compare the objective values of the SVM.
%

%
\begin{proof} For any $\w$, we will have that $p(P\w) \le \tilde{p}(\w)$ since
$\|P\w\|_{\H}^{2} \le \|\w\|_{\H}^{2}$, while the loss term will be identical.
This implies that $p^{*} \le \tilde{p}^{*}$. For the second part of the
inequality, note that:
\begin{align*}
| ( 1-y_i\inner{\w,\bphi(\x_i)} )_{+} - ( 1-y_i\inner{\w,P\bphi(\x_i)} )_{+} | \\
\le | \inner{\w,\bphi(\x_i) - P\bphi(\x_i)} | \\
\le \| \w \|_{\H} \| P^{\perp}\bphi(\x_i) \|_{\H}
\end{align*}
which implies that $\tilde{p}(\w) \le p(\w) + \frac{1}{m}\|\w\|_{\H}
\sum_{i=1}^{m}\|P^{\perp}\bphi( x_i)\|_{\H}$ for any $\w$, and in
particular for $\w^{*}$ the optimum of $p(\w)$. This,
combined with $\|\w^{*}\|_{\H} \le
\frac{1}{\sqrt{\lambda}}$ \citep{ShalevSiSr07} yields:
\begin{align*}
\tilde{p}^{*} & \le p^{*} + \frac{1}{m\sqrt{\lambda}} \sum_{i=1}^{m} \| P^{\perp}\bphi(\x_i) \|_{\H}\\
& = p^{*} + \frac{1}{m\sqrt{\lambda}} \sum_{i=1}^{m} \sqrt{ K( \x_i, \x_i ) - \tilde{K} ( \x_i, \x_i ) }
\end{align*}
\end{proof}

Theorem \ref{thm:primal_objective} suggests using a low-dimensional projection
minimizing $\sum_{i=1}^{m} \sqrt{ K( \x_i, \x_i ) - \tilde{K} ( \x_i, \x_i ) }
= \sum_{i=1}^{m} \| \bphi(\x)-P\bphi(\x) \|$. That is, that one should choose a
subspace of $\H$ with small average distances to the data (not squared
distances as in PCA). The Taylor approximation we suggest is such a projection,
albeit not the optimal one, so we can apply Theorem \ref{thm:primal_objective}
to analyze its approximation properties.

\paragraph{Approximating with Random Features}\label{sec:random_features}

A different option for approximating the mapping $\bphi$ for a radial kernel of
the form $K(\x,\x')=K(\x-\x')$, was proposed by \citet{RahimiRe07}. They
proposed mapping the input data to a randomized low-dimensional feature space
as follows. Let $\hat{K}(\bomega)$ be the real-valued Fourier transform of the kernel
$K(\x-\x')$, namely
\begin{equation}
K(\x-\x')=\int_{\R^d} \hat{K}(\bomega)\cos\,\bomega\!\cdot\!(\x-\x') \, d\bomega
\end{equation}
Bochner's theorem ensures that if $K(\x-\x')$ is
properly scaled, then $\hat{K}(\bomega)$ is a proper probability distribution. Hence:
\begin{align*}
K(\x-\x') & = \E_{\bomega\sim \hat{K}(\bomega)}[\cos\,\bomega\!\cdot\!(\x-\x')] \\ &
= \E_{\bomega\sim \hat{K}(\bomega)}[\cos(\bomega\!\cdot\!\x+\theta) \cdot
\cos(\bomega\!\cdot\!\x'+\theta)].
\end{align*}
The kernel function can then be approximated by independently drawing
$\bomega_1,\ldots,\bomega_D \in \R^d$ from the distribution $\hat{K}(\bomega)$ and
$\theta_1,\ldots,\theta_D$ uniformly from $[0,2\pi]$, and using the explicit
feature mapping:
\begin{equation}
\phia_j(\x)= \cos(\bomega_j\!\cdot\!\x+\theta_j)
\end{equation}
In the case of the Gaussian kernel,
$K(\x-\x')=\exp\left(-\|\x-\x'\|^2/2\sigma^2\right)$, and
$\hat{K}(\bomega)=(2\pi)^{-D/2}\exp\left(-\|\bomega\|^2/2\sigma^2\right)$
defines a Gaussian distribution, from which it is easy to draw i.i.d. samples.

The following guarantee was provided on the convergence of kernel values
$\tilde{K}(\x,\x')=\inner{\bphia(\x),\bphia(\x')}$ corresponding to the random
Fourier feature mapping:
\begin{claim}[Rahimi and Recht, Claim 1]\label{claim:rahimirecht1}
Let $\tilde{K}$ be the kernel defined by $D$ random Fourier features, and $R$
be the radius (in the input space) of the training set, then for any
$\epsilon>0$:
%
\begin{equation}\label{eq:fourier_error}
Pr\left[\sup_{\|x\|,\|y\|\leq R}\left\vert
K\left(x,y\right)-\tilde{K}\left(x,y\right)\right\vert \ge\epsilon\right]
\le
2^{8}\frac{d}{\epsilon^{2}}\left(\frac{R}{\sigma}\right)^{2}e^{-\frac{D\epsilon^{2}}{4\left(2+d\right)}}
\end{equation}
\end{claim}
It is also worth mentioning that the random Fourier features are invariant to
translations and rotations, as is the kernel itself. However, due to the fact
that each corresponds to an independent random projection, a collection of such
features will not, in general, be an orthogonal projection, implying that
Theorem \ref{thm:primal_objective} does not apply.

\section{Taylor Features}\label{sec:taylor_features}

In this section we present an alternative approximation of the Gaussian kernel,
which will be obtained by a projection onto a subspace of $\H$. The idea is to
use the Taylor series expansion of the Gaussian kernel function with respect to
$\inner{\x,\x'}$, where each term in the Taylor series can then be expressed as
a sum of matching monomials in $\x$ and $\x'$. More specifically, we express
the Gaussian kernel as:
\begin{equation}\label{eq:express_Gaussian}
K(\x,\x') = e^{-\frac{\|\x-\x'\|^2}{2\sigma^2}} =
e^{-\frac{\|\x\|^2}{2\sigma^2}}e^{-\frac{\|\x'\|^2}{2\sigma^2}}e^{\frac{\inner{\x,\x'}}{\sigma^2}}
\end{equation}
The first two factors depend on $\x$ and $\x'$ separately, so we focus on the
third factor. The term $z=\inner{\x,\x'}/\sigma^2$ is a real number, and using
the (scalar) Taylor expansion of $e^{z}$ around $z=0$ we have:
\begin{equation}\label{eq:scalar_taylor}
e^{\frac{\inner{\x,\x'}}{\sigma^2}} = \sum_{k=0}^{\infty}
\frac{1}{k!}\left(\frac{\inner{\x,\x'}}{\sigma^2}\right)^k
\end{equation}
We now expand:
\begin{equation}\label{eq:multinomial_exp}
\inner{\x,\x'}^k=(\sum_{i=1}^d \x_i \x'_i)^k=\sum_{j \in [d]^k} \left(
\prod_{i=1}^k \x_{j_i} \right) \left( \prod_{i=1}^k \x'_{j_i} \right)
\end{equation}
where $j$ enumerates over all selections of $k$ coordinates of $\x$ (for
simplicity of presentation, we allow repetitions and enumerate over different
orderings of the same coordinates, thus avoiding explicitly writing down the
multinomial coefficients). We can think of \eqref{eq:multinomial_exp} as an
inner product between degree $k$ monomials of the coordinates of $\x$ and
$\x'$. Plugging this back into \eqref{eq:scalar_taylor} and
\eqref{eq:express_Gaussian} results in the following explicit feature
representation for the Gaussian kernel:
\begin{equation}
\label{eq:taylor_features} \phi_{k,j}\left( \x \right) =
e^{-\frac{\|\x\|^2}{2\sigma^2}} \frac{1}{\sigma^{k}\sqrt{k!}}\prod_{i=0}^{k}
{\x}_{j_{i}}
\end{equation}
with $K(\x,\x') = \inner{\bphi(\x),\bphi(\x')} = \prod_{k=0}^{\infty} \prod_{j
\in [d]^k} \phi_{k,j}(\x)\phi_{k,j}(\x')$. Now, for our approximate feature
space, we project onto the coordinates of $\bphi(\cdot)$ corresponding to
$k\leq r$, for some degree $r$. That is, we take
$\phia_{k,j}(\x)=\phi_{k,j}(\x)$ for $k\leq r$. This corresponds to truncating
the Taylor expansion \eqref{eq:scalar_taylor} after the $r$th term.

We would like to bound the error introduced by this approximation, i.e. bound
$| K(\x,\x')-\tilde{K}(\x,\x')|$ where:
%
\begin{equation}
\label{eq:taylor_kernel_expansion} \tilde{K}(\x,\x')=\inner{\bphia(\x),\bphia(x')}
= e^{-\frac{\|\x\|^2+\|\x'\|^2}{2\sigma^2}} \sum_{k=0}^{r} \frac{1}{k!}\left(\frac{\inner{\x,\x'}}{\sigma^2}\right)^k
\end{equation}
The difference $|K(\x,\x')-\tilde{K}(\x,\x')|$ is given (up to the scaling by
the leading factor) by the higher order terms of the Taylor expansion of $e^z$,
which by Taylor's theorem are bounded by $\frac{z^{r+1}}{(r+1)!}e^\alpha$ for
some $|\alpha|\leq|z|$. We may bound $|\alpha|\leq \inner{\x,\x'}/\sigma^2 $
and $|\inner{\x,\x'}|\leq \|\x\|\,\|\x'\|$, obtaining:
%
\begin{align}
\notag \left\vert K(\x,\x')-\tilde{K}(\x,\x')\right\vert
& \le e^{-\frac{\left\|\x\right\|^2+\left\|\x'\right\|^2}{2\sigma^2}}
\frac{1}{\left(r+1\right)!} \left(\frac{\left\langle \x,\x'
\right\rangle}{\sigma^2}\right)^{r+1} e^{\frac{\left\langle \x,\x'
\right\rangle}{\sigma^2}} \\
%
%
& \label{eq:taylor_error} \le \frac{1}{(r+1)!} \left(
\frac{\|\x\|\,\|\x'\|}{\sigma^2} \right)^{r+1}
\end{align}
As for the dimensionality $D$ of $\bphia(\cdot)$ (i.e. the number of features
of degree not more than $r$), as presented we have $d^k$ features of degree
$k$. But this ignores the fact that many features are just duplicates resulting
from different permutations of $j$. Collecting these into a single feature for
each distinct monomial (with the appropriate multinomial coefficient), we have
${{d+k-1} \choose {k}}$ features of degree $k$, and a total of $D={{d+r}
\choose {r}}$ features of degree at most $r$.

%
%
%
%
%

\section{Theoretical Comparison of Taylor and Random Fourier Features}\label{sec:comparison}

We now compare the error bound of the Taylor features given in
\eqref{eq:taylor_error} to the probabilistic bound of the random Fourier
features given in \eqref{eq:fourier_error}.

We first note that each Taylor feature may be calculated in {\em constant}
time, because each degree-$k$ feature may be derived from a degree-$k \!- \! 1$
feature by multiplying it by a constant times an element of $\x$. In fact,
because each feature is proportional to a product of elements of $\x$, on
sparse datasets, the Taylor features will themselves be highly sparse, enabling
one to entirely avoid calculating many features. For a vector $\x$ with
$\tilde{d}$ nonzeros, one may verify that there will be ${{\tilde{d}+r} \choose
{r}} = O(\tilde{d}^{r})$ nonzero features of degree at most $r$, which can all
be computed in overall time $O(\tilde{d}^{r})$.

In contrast, computing each Fourier feature requires $O(\tilde{d})$ time on a
vector with $\tilde{d}$ nonzeros, yielding an overall time of $O(D \cdot
\tilde{d})$ to compute $D$ random Fourier features.

With this in mind, we will define $B$ as a ``budget'' of operations, and will
take as many features as may be computed within this budget, assuming that each
nonzero Taylor feature may be calculated in one operation, and each Fourier
feature in $\tilde{d}$. Setting $\delta = Pr\left[\left\vert
K(x,y)-\tilde{K}(x,y)\right\vert \ge\epsilon\right]$ and solving
\eqref{eq:fourier_error} for $\epsilon$, with $D \approx \frac{B}{\tilde{d}}$,
yields that with probability $1-\delta$, for the Fourier features:
\begin{align}\label{eq:FourierKK}
\left\vert K(x,x')-\tilde{K}(x,x')\right\vert & \approx O\left( \sqrt{
	\frac{\tilde{d}d}{B} \log\left(\frac{d R^2}{\delta \sigma^{2}}\right) }
	\right)
\end{align}

For the Taylor features, $B = {{\tilde{d}+r} \choose {r}}$ implies that $r+1
\gtrapprox \frac{\log B}{\log \tilde{d}}$. Applying Stirling's approximation to
\eqref{eq:taylor_error} yields:
%
\begin{equation}
\label{eq:TaylorKK} \left \vert K(x,x')-\tilde{K}(x,x')\right\vert
\approx O\left( \sqrt{\frac{\log\tilde{d}}{\log B}} B^{- \left(\frac{1}{\log\tilde{d}} \log \left( \frac{\sigma^2 \log B}{R^2 \log \tilde{d}} \right)\right)}
\right)
\end{equation}

Neither of the above bounds clearly dominates the other. The main advantage of
the Taylor approximation, also seen in the above bounds, is that its
performance only depends on the number of non-zero input dimensions
$\tilde{d}$, unlike the Fourier random features which have a cost which scales
quadratically with the dimension, and even for sparse data will depend
(linearly) on the overall dimensionality. The computational budget required for
the Taylor approximation is polynomial in the number of {\em non-zero}
dimensions, but exponential in the effective radius $(R/\sigma)$. Once the
budget is high enough, however, these features can yield a polynomial decrease
in the approximation error. This suggest that Taylor approximation is
particularly appropriate for sparse (potentially high-dimensional) data sets
with a moderate number of non-zeros, and where the kernel bandwidth is on the
same order as the radius of the data (as is often the case).

%
%
%
%
%
%
%
%

\section{Experiments}\label{sec:experiments}

\begin{table*}[tb]

\caption{
\small Datasets used in our experiments. The ``Dim'' and ``NZ'' columns
contain the total number of elements in each training/testing vector, and the
average number of nonzeros elements, respectively.
}

\begin{centering}

\begin{tabular}{ccccc|ccc|cc}
\hline
\multicolumn{5}{c|}{Dataset} & \multicolumn{3}{c|}{Kernel SVM} & \multicolumn{2}{c}{Linear SVM}\\
Name & Train size & Test size & Dim & NZ & $C$ & $\sigma^{2}$ & Test error & $C$ & Test error\\
\hline
Adult & 32562  & 16282 & 123 & 13.9 & 1 & 40    & 14.9\% & 8 & 15.0\% \\
Cov1  & 522911 & 58101 & 54  & 12   & 3 & 0.125 & 6.2\%  & 4 & 22.7\% \\
MNIST & 60000  & 10000 & 768 & 150  & 1 & 100   & 0.57\% & 2 & 5.2\%  \\
TIMIT & 63881  & 22257 & 39  & 39   & 1 & 80    & 11.5\% & 4 & 22.7\% \\
\hline
\end{tabular}

\end{centering}

\label{tab:datasets}

\end{table*}

In this section, we describe an empirical comparison of the random Fourier
features of Rahimi and Recht and the polynomial Taylor-based features described
in Section \ref{sec:taylor_features}. The question we ask is: which explicit
feature representation provides a better approximation to the Gaussian kernel,
with a fixed computational budget?

%
Experiments were performed on four datasets, summarized in Table
\ref{tab:datasets}. Adult and MNIST were downloaded from L\'{e}on Bottou's
LaSVM web page. They, along with Blackard and Dean's forest covertype-1 dataset
(available in the UCI machine learning repository), are well-known SVM
benchmark datasets. TIMIT is a phonetically transcribed speech corpus, of which
we use a subset for framewise classification of the stop consonants. From each
10 ms frame of speech, we extracted MFCC features and their first and second
derivatives. Both MNIST and TIMIT are multiclass classification problems, which
we converted into binary problems by performing one-versus-rest, with the digit
$8$ and phoneme \texttt{/k/} being the positive classes, respectively.
The regularization and Gaussian kernel parameters for Adult, MNIST and Cov1 are
taken from \citet{ShalevSiSrCo10}, and are in turn based on those of
\citet{Platt98} and \citet{BordesErWeBo05}. The parameters for TIMIT were found
by optimizing the test error on a held-out validation set.
Of these datasets, all except MNIST are fairly low-dimensional, and all except
TIMIT are sparse. To get a rough sense of the benefit of the Gaussian kernel
for these data sets, we also include in Table \ref{tab:datasets} the best test
error obtained using a linear kernel over $C$s in the range $\left[ 2^{-6}, 2^6
\right]$.

\begin{figure*}[tb]

\noindent \begin{centering}
\begin{tabular}{ @{} L @{} T @{} T @{} T @{} }

& Adult & MNIST & TIMIT \\
\rotatebox{90}{\scriptsize{Primal objective}} &
\includegraphics[width=0.29\textwidth]{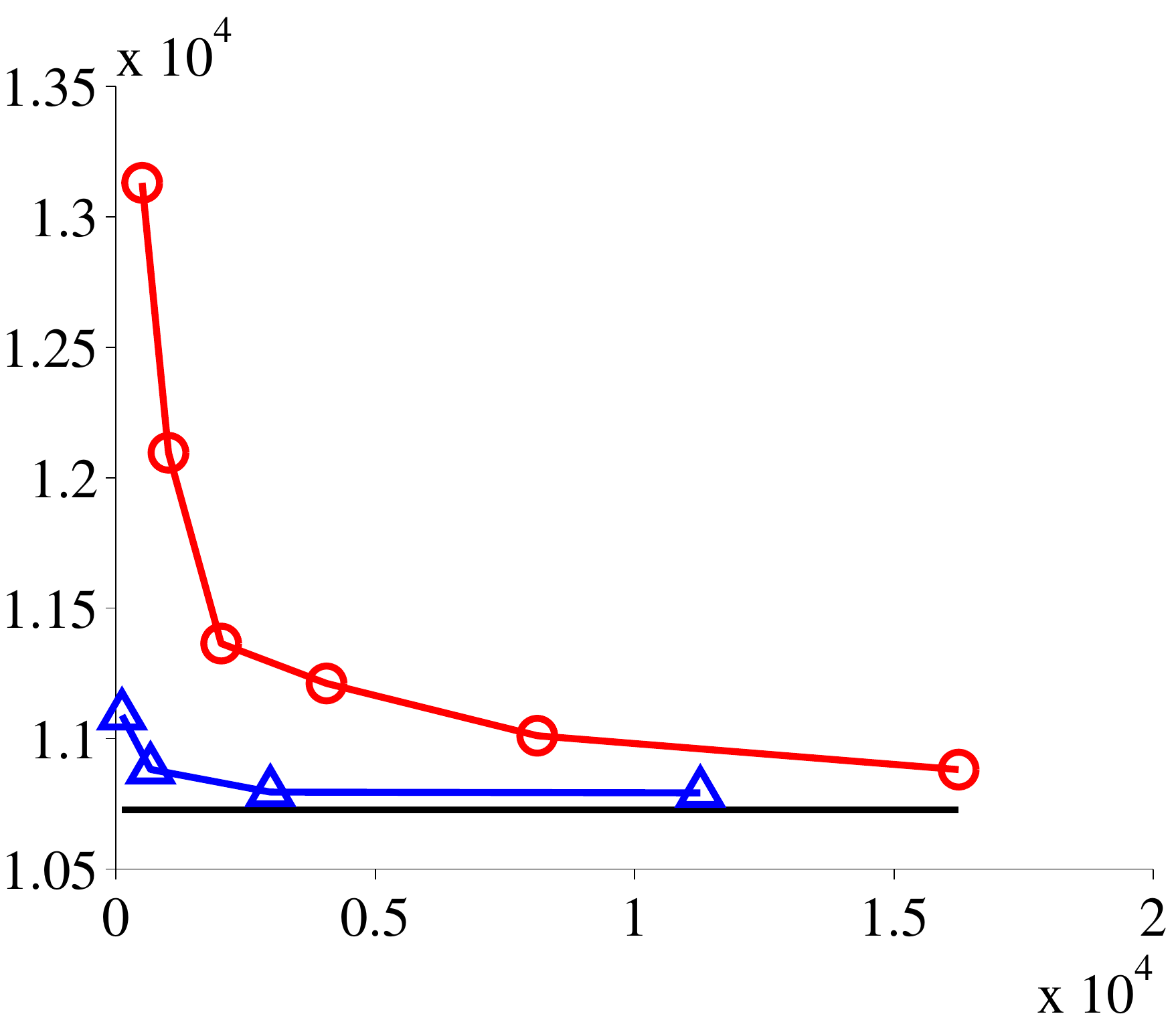} &
\includegraphics[width=0.29\textwidth]{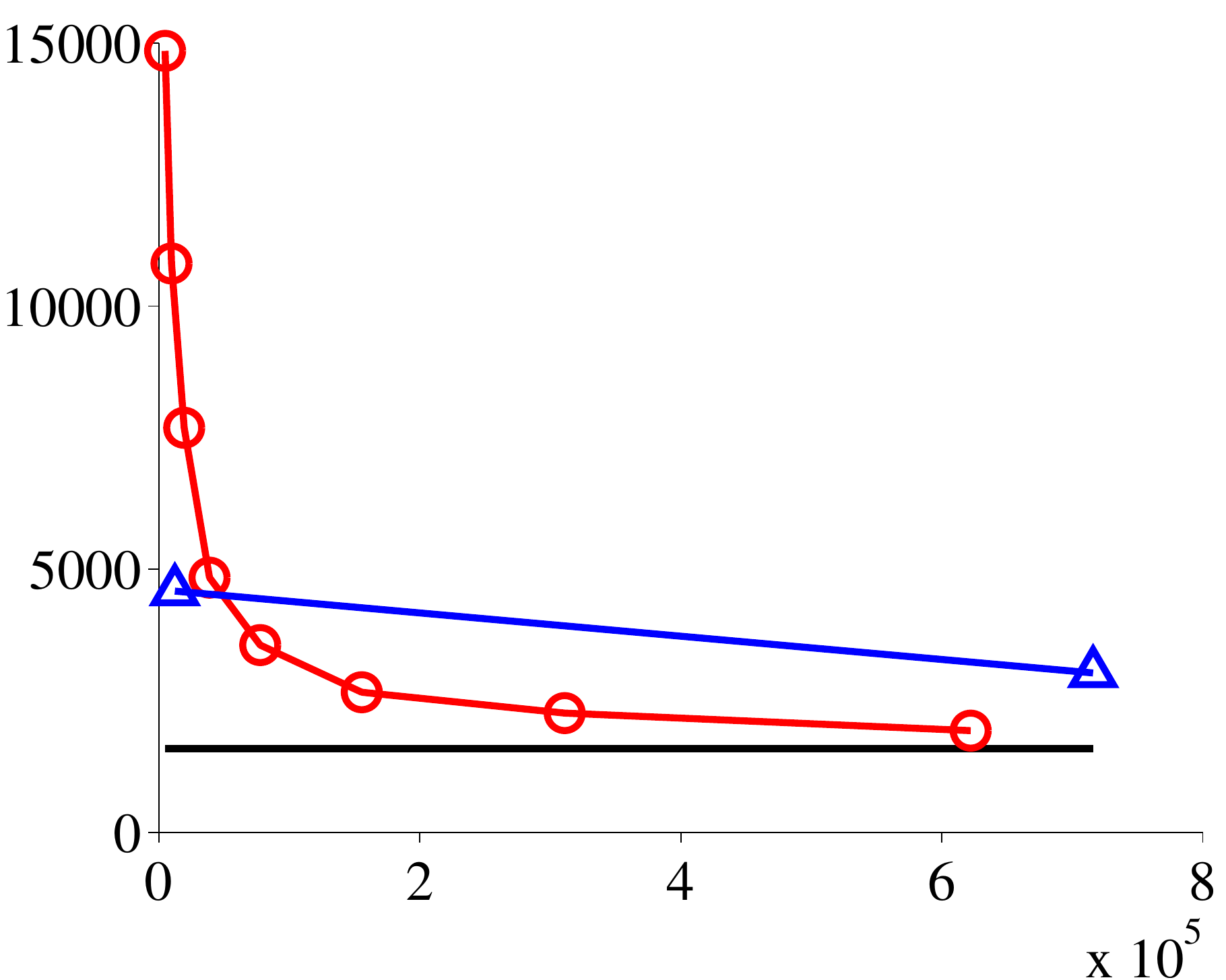} &
\includegraphics[width=0.29\textwidth]{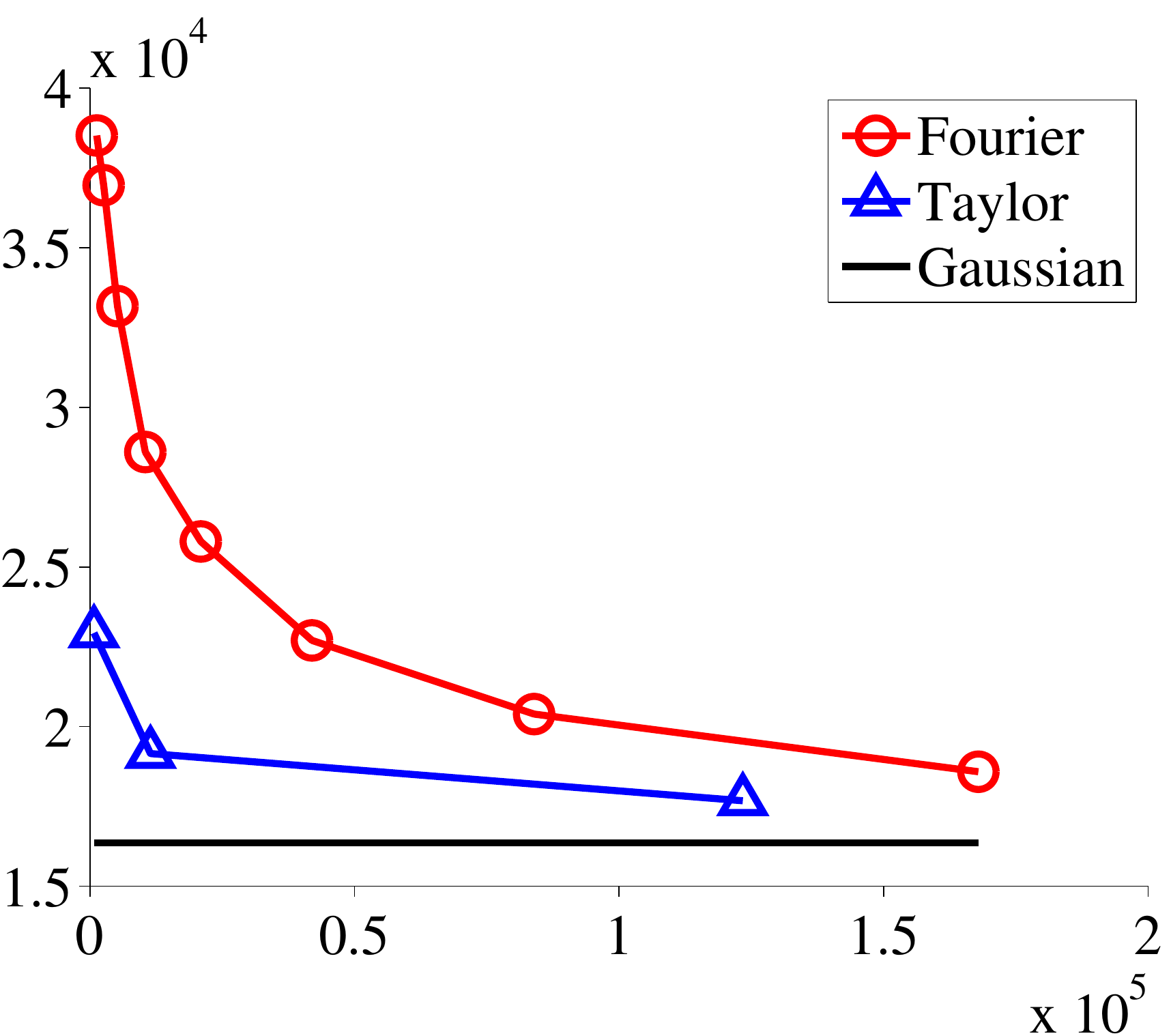} \\
\rotatebox{90}{\scriptsize{Testing classification error}} &
\includegraphics[width=0.29\textwidth]{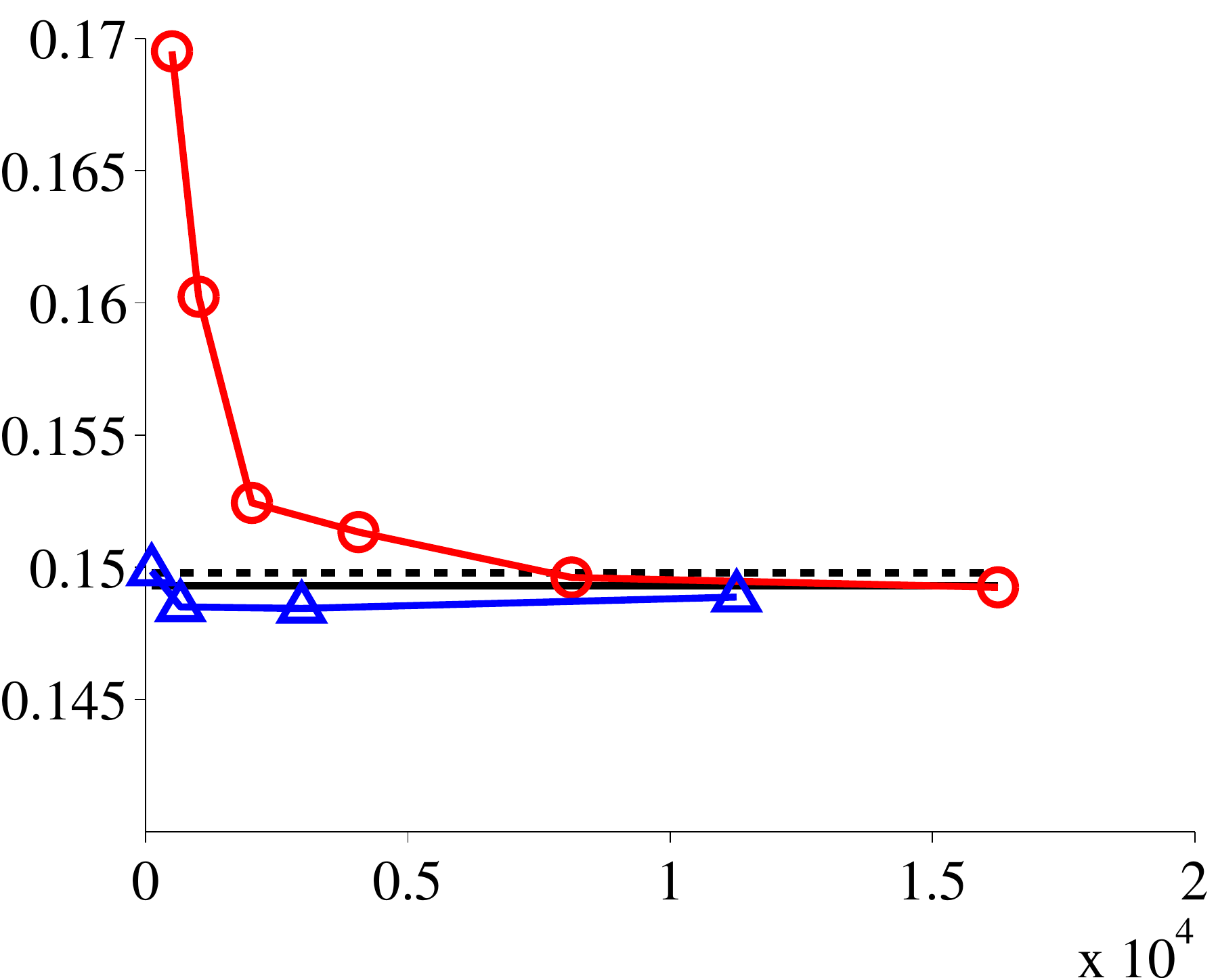} &
\includegraphics[width=0.29\textwidth]{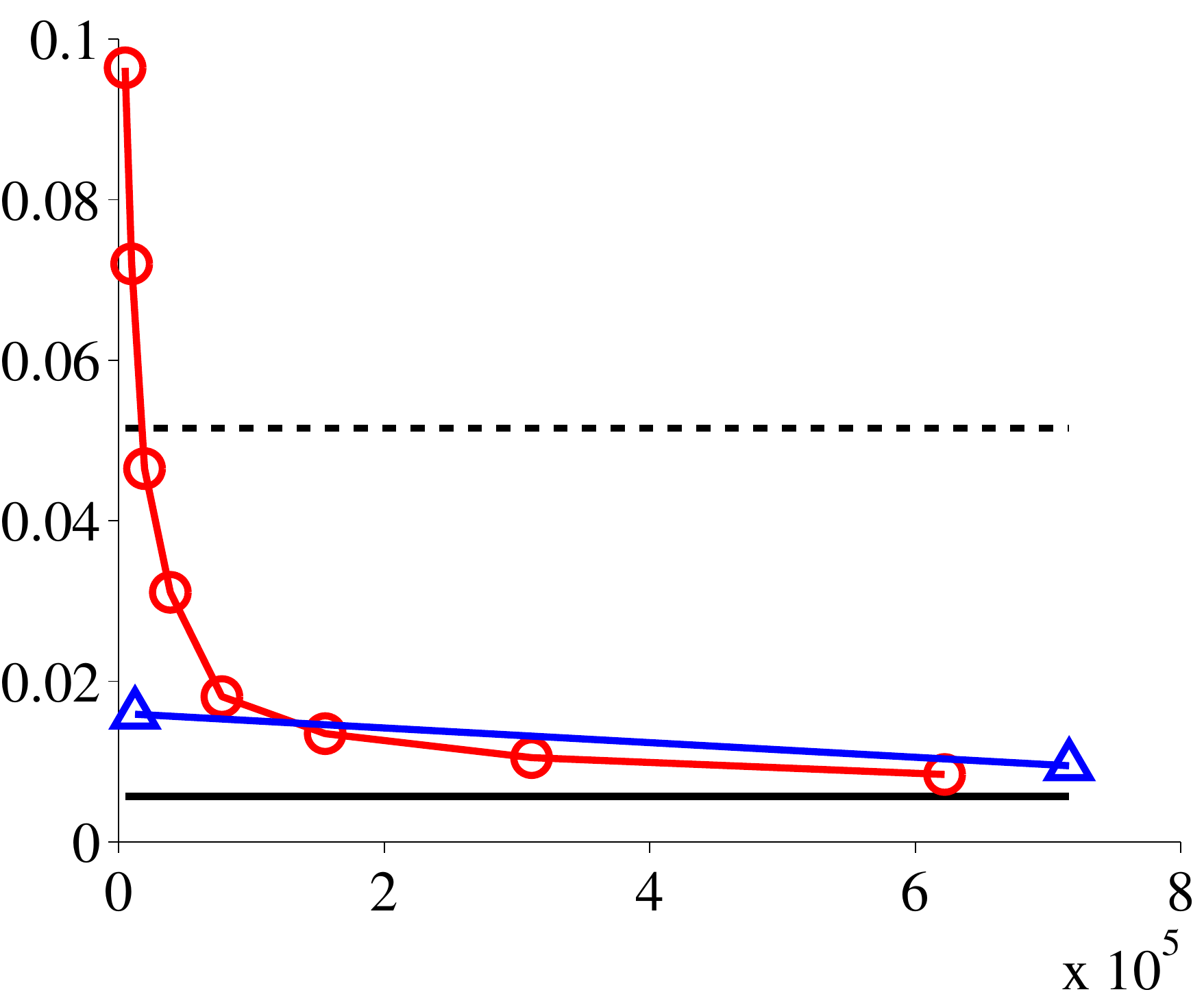} &
\includegraphics[width=0.29\textwidth]{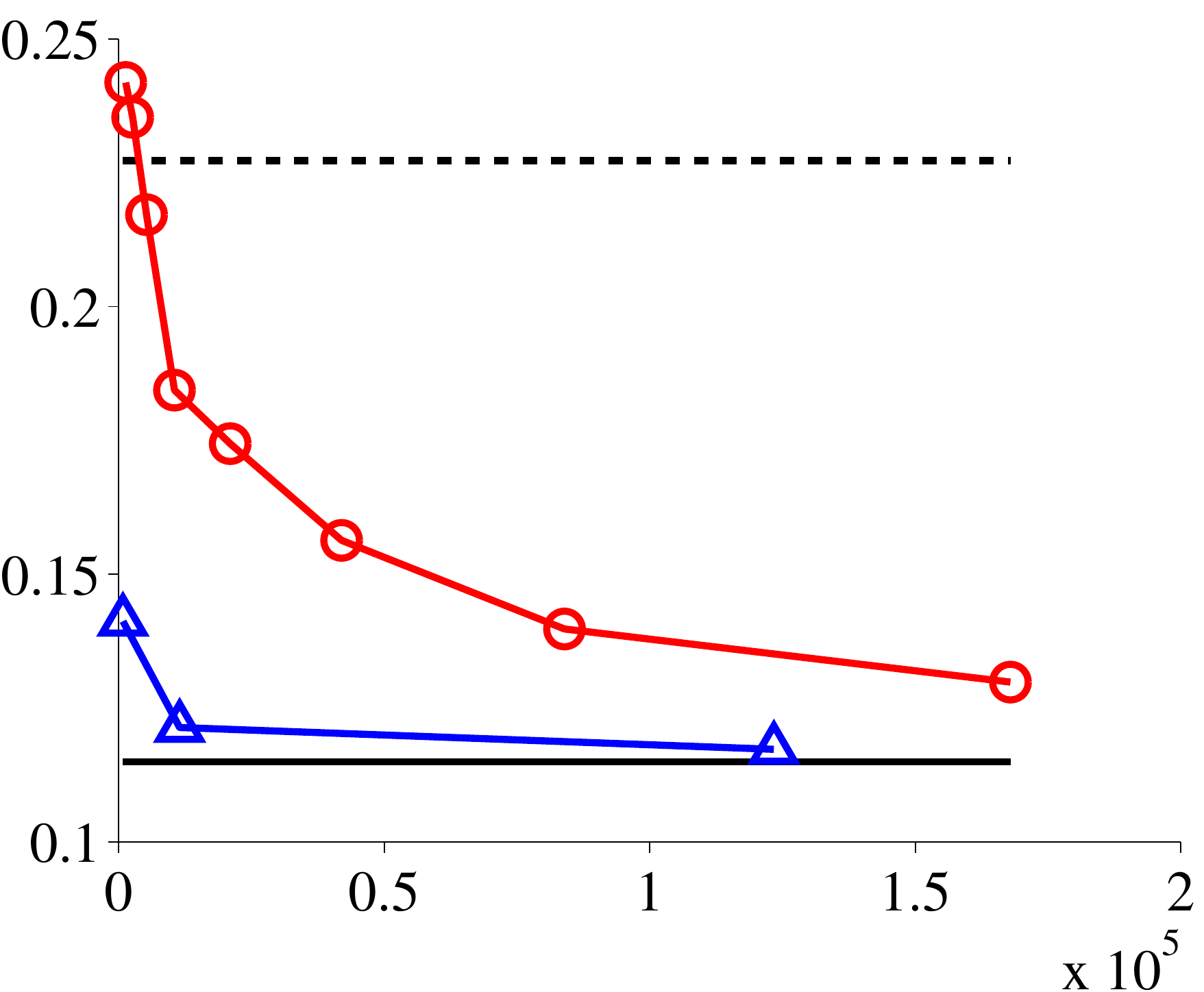} \\[-10pt]
& \scriptsize{Computational cost} & \scriptsize{Computational cost} & \scriptsize{Computational cost} \\

\end{tabular}
\end{centering}

\caption{
Primal objectives and testing classification errors for various numbers of
Fourier and Taylor features. For the Fourier features, the markers correspond
to numbers of features which are powers of two, starting at $32$. For Taylor,
each marker corresponds to a degree, starting at $2$. The cost of calculating
$\tilde{\phi}$, in units of floating point operations, is displayed on the
horizontal axis. The solid black lines are the primal objective function value
and testing classification error achieved by the optimal solution to the
Gaussian kernel SVM problem, while the dashed lines in the bottom plots are the
testing classification error achieved by a linear SVM.
}

\label{fig:experiments}

\end{figure*}

\begin{figure*}[tb]

\noindent \begin{centering}
\begin{tabular}{ @{} L @{} S @{} L @{} S @{} L @{} S @{} }

\multicolumn{6}{c}{Cov1} \\

\rotatebox{90}{\scriptsize{Primal objective}} &
\includegraphics[width=0.29\textwidth]{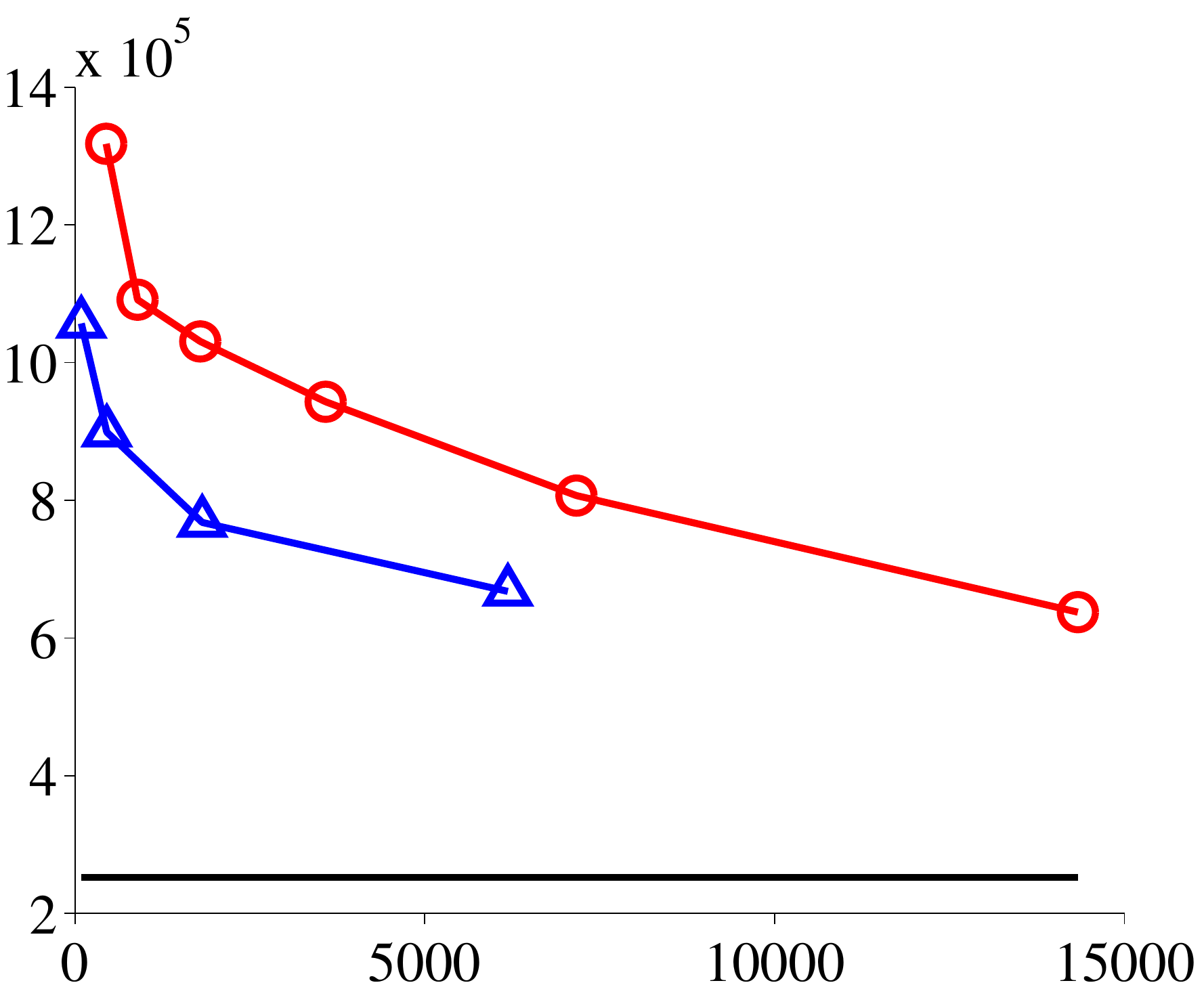} &
\rotatebox{90}{\scriptsize{Primal objective}} &
\includegraphics[width=0.29\textwidth]{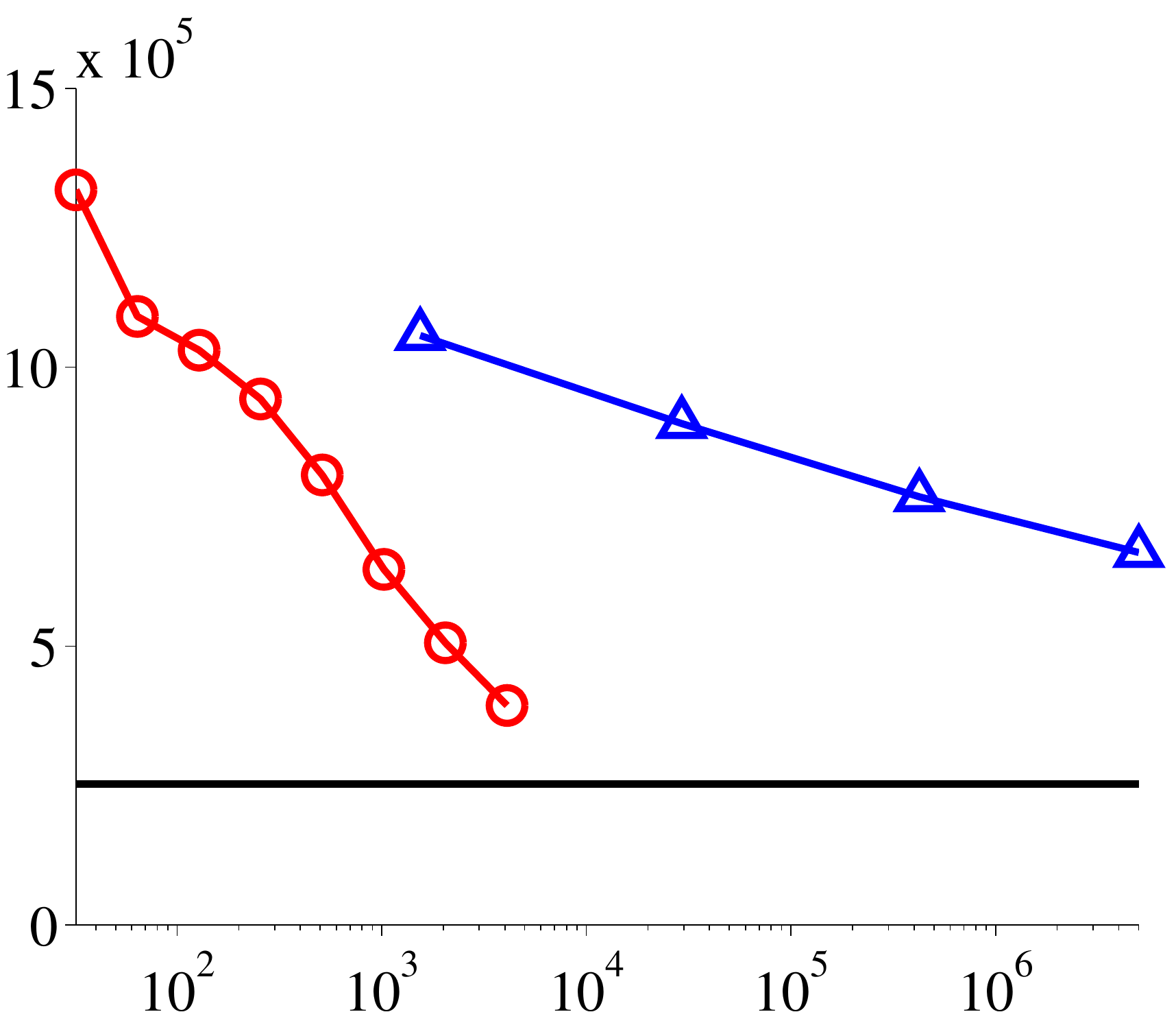} &
\rotatebox{90}{\scriptsize{Average $\left\vert K - \tilde{K} \right\vert$}} &
\includegraphics[width=0.29\textwidth]{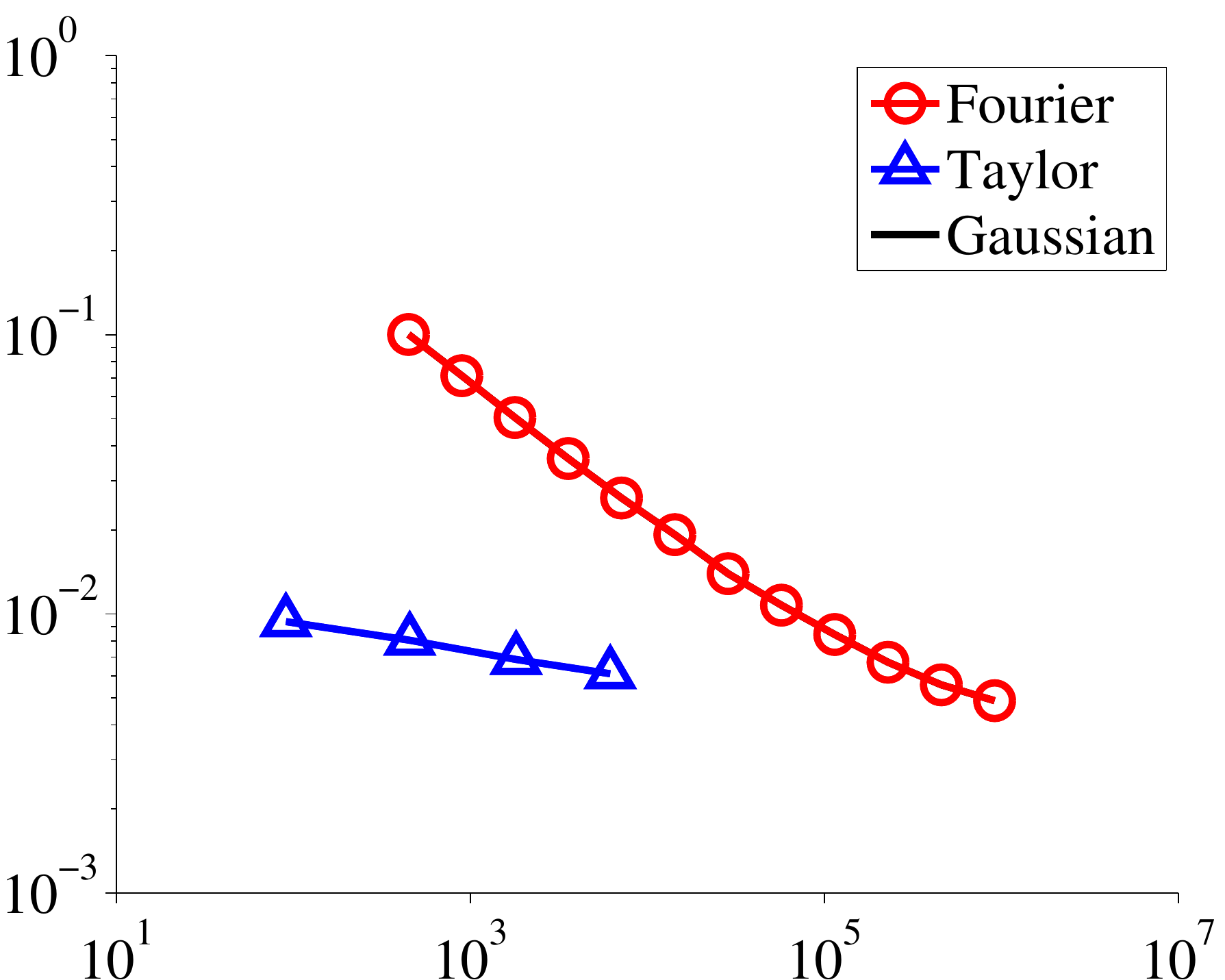} \\[-3pt]
& \scriptsize{Computational cost} & & \scriptsize{\# features} & & \scriptsize{Computational cost} \\

\rotatebox{90}{\scriptsize{Testing classification error}} &
\includegraphics[width=0.29\textwidth]{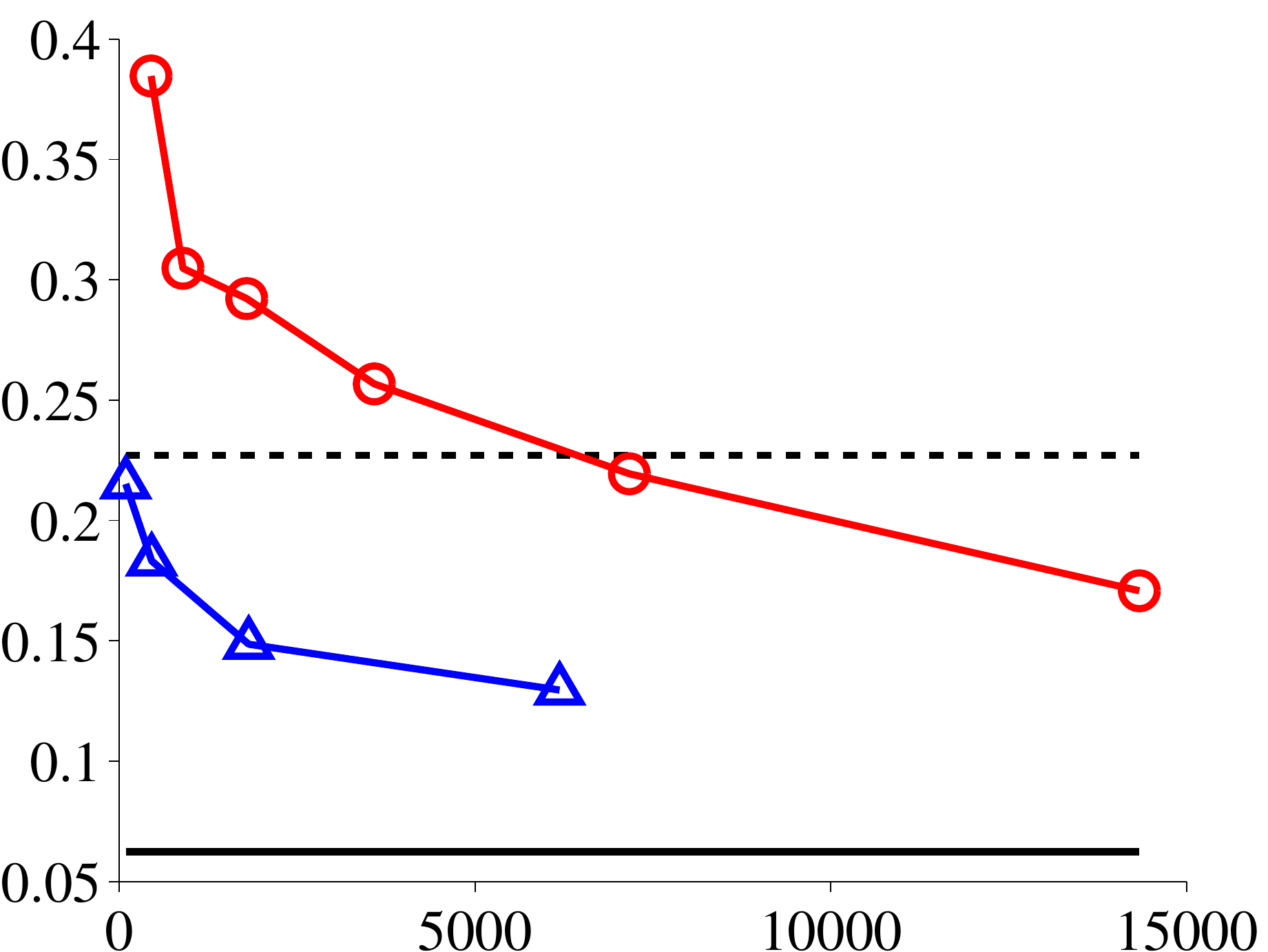} &
\rotatebox{90}{\scriptsize{Testing classification error}} &
\includegraphics[width=0.29\textwidth]{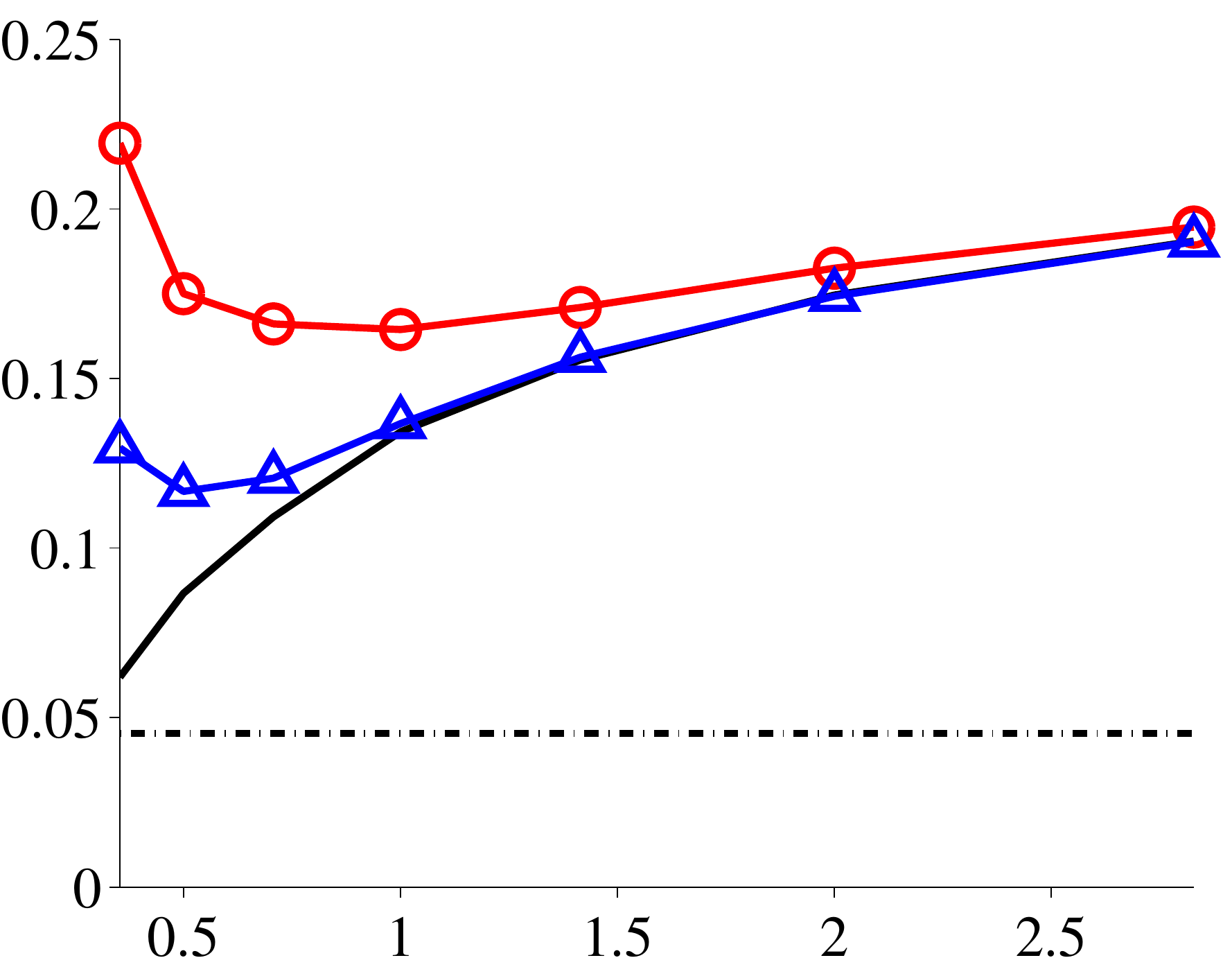} &
\rotatebox{90}{\scriptsize{Average $\left\vert K - \tilde{K} \right\vert$}} &
\includegraphics[width=0.29\textwidth]{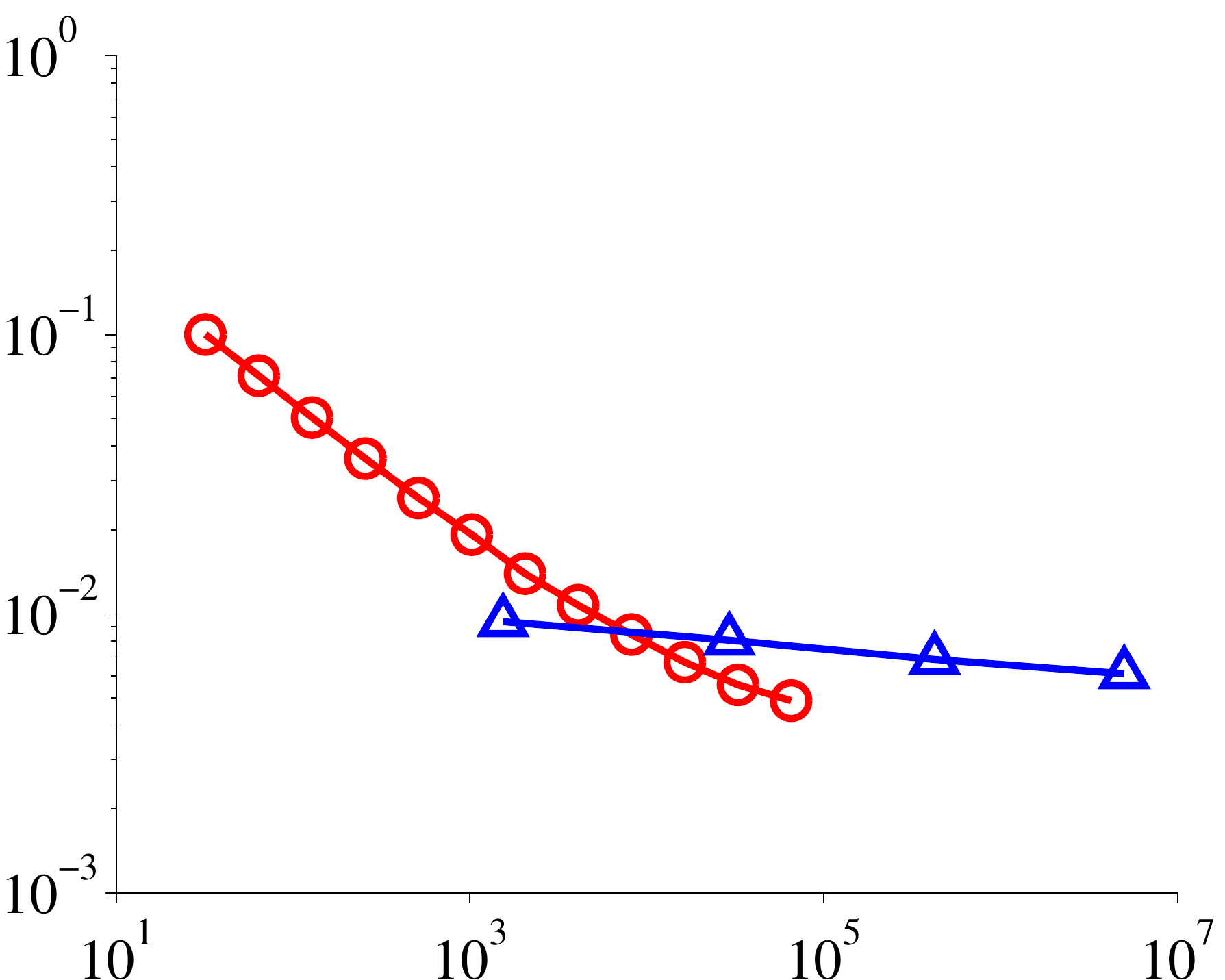} \\[-3pt]
& \scriptsize{Computational cost} & & \scriptsize{$\sigma$} & & \scriptsize{\# features} \\

\end{tabular}
\end{centering}

\caption{
Left column: same as Figure \ref{fig:experiments}. Top middle: primal objective
as a function of the total number of features (in log scale). Bottom middle:
test error as a function of the Gaussian kernel parameter $\sigma$, for Taylor
and random Fourier expansions of the same computational cost, compared with the
true Gaussian kernel. The dot-dashed line indicates the performance of a 1-NN
classifier, trained using the ANN library \citep{AryaMo93,MountAr06,Bagon09}.
Right column: average value of the approximation error $| K\left( x, x' \right)
- \tilde{K}\left( x, x' \right) |$ over 100000 randomly chosen pairs of
training vectors, in terms of both computational cost and total number of
features.
}

\label{fig:cov1}

\end{figure*}

For each of the data sets, we compared the value of the (primal) SVM objective
and the classification performance (on the test set) achieved using varying
numbers of Taylor and Fourier features. Results are reported in Figure
\ref{fig:experiments} and in the left column of Figure \ref{fig:cov1}. We
report results in units of the number of floating-point operations required to
calculate each feature vector, taking into account sparsity, as discussed in
Section \ref{sec:comparison}. As was discussed earlier, this is the dominant
cost in stochastic optimization methods such as Pegasos \citep{ShalevSiSr07} and
stochastic dual coordinate ascent \citep{HsiehChLiKeSu08}, which are the fastest
methods of training large-scale liner SVMs. We used a fairly optimized SGD
implementation into which the explicit feature vector calculations were
integrated. Our actual runtimes are indeed in line with the calculated
computational costs (we prefer reporting the theoretical cost as it is not
implementation or architecture dependent).

As can be seen from Figures \ref{fig:experiments} and \ref{fig:cov1}, the
computational cost required to obtain the same SVM performance is typically
{\em lower} when using the Taylor features than the Fourier features, despite
the exponential growth of the number of features as a function of the degree.
The exception is the MNIST dataset, which has a fairly high number (over 150)
of non-zero dimensions per data point, yielding an extremely sharp increase in
the computational costs of higher-degree Taylor feature expansions.

To better appreciate the difference between the dependence on the {\em number}
of features and that on the {\em computational cost} we include more detailed
results for the Cov1 dataset, in Figure \ref{fig:cov1}. Here we again plot the
value of the SVM objective, this time both as function of the number of
features and as a function of the computational cost. As expected, the Fourier
features perform much better as a function of the number of features, but, as
argued earlier, we should be more concerned with the cost of calculating them.
In order to directly measure how well each feature representation approximates
the Gaussian kernel, we also include in Figure \ref{fig:cov1} a comparison of
the average approximation error.

Next, we consider the effect of the bandwidth parameter $\sigma$ on the Taylor
and Fourier approximations---note that the theoretical analysis for both
methods deteriorates significantly when the bandwidth decreases. This is
verified in the bottom-middle plot of Figure \ref{fig:cov1}, which shows that
the (test) classification error of the two approximations (with the same fixed
computational budget) deteriorates, relative to that of the true Gaussian SVM,
as the bandwidth decreases. This deterioration can be observed on other data
sets as well. On Cov1, the deterioration is so strong that even though the
generalization performance of the true Gaussian Kernel SVM keeps improving as
the bandwidth decreases, the test errors for both approximations actually start
increasing fairly quickly. It should be noted that Cov1 is atypical in this
regard: nearest-neighbor classification achieves almost optimal results on this
dataset (the dot-dashed line in the bottom-middle plot of Figure
\ref{fig:cov1}), and so decreasing the bandwidth, which approximates the
nearest-neighbor classifier, is beneficial. In contrast, on the data sets in
Figure \ref{fig:experiments}, the optimal bandwidth for the Gaussian kernel is
large enough to allow good approximation by the Fourier and Taylor
approximations.

Finally, in order to get some perspective on the real-world benefit of the
Taylor features, we also report actual runtimes for a large scale realistic
example. We compared training times for the Gaussian kernel and the Taylor
features, on the full TIMIT dataset, where the goal was framewise phoneme
classification, i.e., given a 10 ms frame of speech the goal is to predict the
uttered phoneme from a set of 39 phoneme symbols. We used the standard split of
the dataset to training, validation test sets, and extracted MFCC features.
With this set of acoustic features the common practice in to use the Gaussian
kernel. Its bandwidth was selected on the validation set to be $\sigma^2=19$.
The training set includes 1.1 million examples, and existing SVM libraries such
as SVMLIB or SVMLight failed to converge in a reasonable amount of time (see
the training time in \citet{SalomonKiOs02}). Using our own implementation with
the exact Gaussian kernel and stochastic dual coordinate ascent, the training
took 313 hours (almost two weeks) on 2GHz Intel Core 2 (using one core). Using
the same implementation with the kernel function replaced by its degree-$3$
Taylor approximation, the training took only 53 hours. The results were almost
the same: multiclass accuracy of $69.6\%$ for the approximated kernel and
$69.8\%$ for the Gaussian kernel. These are state-of-the-art results for this
dataset \citep{SalomonKiOs02,GravesSc05}.

\section{Relationship to the Polynomial Kernel}\label{sec:polynomial}

Like the Taylor feature representation of the Gaussian kernel, the standard
polynomial kernel of degree $r$:
\begin{equation}\label{eq:polyK}
K\left( x, x' \right) = \left( \left\langle x, x' \right\rangle + c \right)^r
\end{equation}
corresponds to a feature space containing all monomials of degree at most $r$.
More specifically, the features corresponding to the kernel \eqref{eq:polyK}
can be written as:
\begin{equation}
\label{eq:polynomial_features} \phi_{k,j} \left( x \right) = \sqrt{{r \choose
k} c^{r-k}} \prod_{i=1}^k x_{j_{i}}
\end{equation}
where, as in \eqref{eq:taylor_features}, $k=0,\ldots,r$ and $j\in\left[ d
\right]^k$ enumerates over all selections of $k$ coordinates in $\x$. The
difference, relative to the Taylor approximation to the Gaussian, is only in a
per-example overall scaling factor based on $\|\x\|$, and in a different
per-degree factor (which depends only on the degree $k$). This weighting by a
degree-dependent factor should not be taken lightly, as it affects
regularization, which is key to SVM training--features scaled by a larger
factor are ``cheaper'' to use, compared to those scaled by a very small factor.
Comparing the degree-dependent scaling in the two feature representations
(\eqref{eq:taylor_features} and \eqref{eq:polynomial_features}), we observe
that the higher degrees are scaled by a much smaller factor in the Taylor
features, owing to the rapidly decreasing dependence on $1/\sqrt{k!}$. This
means that higher degree monomials are comparatively much more expensive for
use in the Taylor features, and that the learned predictor likely relies more
on lower degree monomials.

\begin{table*}[tb]

\caption{
Comparison of the Gaussian kernel and its Taylor approximation to the
polynomial kernel $K\left( x, y \right)=\left(\left\langle x, y \right\rangle +
1 \right)^d$, after scaling the data to have unit average squared norm. Here,
$d$ is the degree of the polynomial. The reported test errors are the minima
over parameter choices taken from a coarse power-of-two based grid, within
which the reported parameters are well inside the interior.  Gaussian kernel
SVMs were optimized using our GPU optimizer, while the others were optimized by
running Pegasos for $100$ epochs.
}

\begin{centering}

\begin{tabular}{c|ccc|cccc|ccc}
\hline
\multicolumn{1}{c|}{} & \multicolumn{3}{c|}{Gaussian} & \multicolumn{4}{c|}{Taylor} & \multicolumn{3}{c}{Polynomial}\\
Dataset & $C$ & $\sigma^2 $ & Test error & degree & $C$ & $\sigma^2 $ & Test error & degree & $C$ & Test error \\
\hline
Adult & 4  & 100     & 14.9\% & 4 & 8    & 200 & 14.7\% & 4 & 4   & 14.8\% \\
MNIST & 8  & 100     & 0.42\% & 2 & 2048 & 200 & 0.54\% & 2 & 256 & 0.58\% \\
TIMIT & 2  & 40      & 10.8\% & 3 & 8    & 200 & 11.4\% & 3 & 64  & 11.6\% \\
Cov1  & 16 & 0.03125 & 3.3\%  & 4 & 128  & 0.5 & 12.3\% & 4 & 512 & 13.6\% \\
\hline
\end{tabular}

\end{centering}

\label{tab:polynomial}

\end{table*}

Nevertheless, the space of allowed predictors is nearly the same with both
types of features, raising the question of how strong the actual effect of the
different per-degree weighting is. The fact that all of the features in the
Taylor representation are scaled by a factor depending on $\|x\|$ should make
little difference on many datasets, as it affects all of the features of a
given example equally. Likewise, if most of the {\em used} features are of the
same degree, then we could perhaps correct for the degree-based scaling by
changing the regularization parameter. The problem, of course, is searching for
and selecting this parameter.

We checked if we could find a substantial difference in performance between the
Taylor and standard polynomial features. Because the dependence on the
regularization parameter necessitated a search over the parameter space, we
conducted a rough experiment in which we tried different parameters, and
compared the best error achieved on the test set using the true Gaussian
kernel, a Taylor approximation, and a standard polynomial kernel of the same
degree. The results are summarized in Table \ref{tab:polynomial}. These
experiments indicate that the standard polynomial features might be sufficient
for approximating the Gaussian. Still, the Taylor features are just as easy to
compute and use, and have the advantage that they use the same parameters as
the Gaussian kernel. Hence, if we already have a sense of good bandwidth and
$C$ parameters for the Gaussian kernel, we can use the same values for the
Taylor approximation.

\section{Summary}\label{sec:conclusion}

The use of explicit monomial features of the form of
\eqref{eq:polynomial_features} has been discussed recently as a way of speeding
up training with the polynomial kernel \citep{SonnenburgFr10,ChangHsChRiLi10}.
Our analysis and experiments indicate that a similar monomial representation is
also suitable for approximating the Gaussian kernel. We argue that such
features might often be preferable to the random Fourier features recently
suggested by \citet{RahimiRe07}. This is especially true on sparse datasets
with a moderate number (up to several dozen) of non-zero dimensions per data
point.

Although we have only focused on binary classification, it is important to note
that the this explicit feature representation can be used anywhere else
$\ell_2$ regularization is used. This includes multiclass, structured and
latent SVMs. The use of such feature expansions might be particularly
beneficial to structured SVMs, since these problems are hard to solve with only
a kernel representation.

\bibliographystyle{abbrvnat}
\bibliography{main}

\begin{thebibliography}{18}
\providecommand{\natexlab}[1]{#1}
\providecommand{\url}[1]{\texttt{#1}}
\expandafter\ifx\csname urlstyle\endcsname\relax
  \providecommand{\doi}[1]{doi: #1}\else
  \providecommand{\doi}{doi: \begingroup \urlstyle{rm}\Url}\fi

\bibitem[Arya and Mount(1993)]{AryaMo93}
S.~Arya and D.~M. Mount.
\newblock Approximate nearest neighbor queries in fixed dimensions.
\newblock In \emph{Proc. SODA 1993}, pages 271--280, 1993.

\bibitem[Bagon(2009)]{Bagon09}
S.~Bagon.
\newblock Matlab class for {ANN}, February 2009.
\newblock URL \url{http://www.wisdom.weizmann.ac.il/~bagon/matlab.html}.

\bibitem[Balcan et~al.(2006)Balcan, Blum, and Vempala]{BalcanBlVe06}
M.-F. Balcan, A.~Blum, and S.~Vempala.
\newblock Kernels as features: On kernels, margins, and low-dimensional
  mappings.
\newblock In \emph{Machine Learning}, volume 65~(1), pages 79--94. Springer
  Netherlands, October 2006.

\bibitem[Bordes et~al.(2005)Bordes, Ertekin, Weston, and
  Bottou]{BordesErWeBo05}
A.~Bordes, S.~Ertekin, J.~Weston, and L.~Bottou.
\newblock Fast kernel classifiers with online and active learning.
\newblock \emph{JMLR}, 6:\penalty0 1579--1619, September 2005.

\bibitem[Chang et~al.(2010)Chang, Hsieh, Chang, Ringgaard, and
  Lin]{ChangHsChRiLi10}
Y.-W. Chang, C.-J. Hsieh, K.-W. Chang, M.~Ringgaard, and C.-J. Lin.
\newblock Training and testing low-degree polynomial data mappings via linear
  {SVM}.
\newblock \emph{JMLR}, 99:\penalty0 1471--1490, August 2010.

\bibitem[Fine and Scheinberg(2002)]{FineSc02}
S.~Fine and K.~Scheinberg.
\newblock Efficient {SVM} training using low-rank kernel representations.
\newblock \emph{JMLR}, 2:\penalty0 243 -- 264, March 2002.

\bibitem[Graves and Schmidhuber(2005)]{GravesSc05}
A.~Graves and J.~Schmidhuber.
\newblock Framewise phoneme classification with bidirectional {LSTM} and other
  neural network architectures.
\newblock \emph{Neural Networks}, 18:\penalty0 602--610, 2005.

\bibitem[Hsieh et~al.(2008)Hsieh, Chang, Lin, Keerthi, and
  Sundararajan]{HsiehChLiKeSu08}
C.-J. Hsieh, K.-W. Chang, C.-J. Lin, S.~S. Keerthi, and S.~Sundararajan.
\newblock A dual coordinate descent method for large-scale linear {SVM}.
\newblock In \emph{Proc. ICML 2008}, pages 408--415, 2008.

\bibitem[Mount and Arya(2006)]{MountAr06}
D.~M. Mount and S.~Arya.
\newblock {ANN}: A library for approximate nearest neighbor searching, August
  2006.
\newblock URL \url{http://www.cs.umd.edu/~mount/ANN}.

\bibitem[Platt(1998)]{Platt98}
J.~C. Platt.
\newblock Fast training of support vector machines using {S}equential {M}inimal
  {O}ptimization.
\newblock In B.~Sch\"olkopf, C.~Burges, and A.~Smola, editors, \emph{Advances
  in Kernel Methods - Support Vector Learning}. MIT Press, 1998.

\bibitem[Rahimi and Recht(2007)]{RahimiRe07}
A.~Rahimi and B.~Recht.
\newblock Random features for large-scale kernel machines.
\newblock In \emph{Proc. NIPS 2007}, 2007.

\bibitem[Rahimi and Recht(2008)]{RahimiRe08}
A.~Rahimi and B.~Recht.
\newblock Uniform approximation of functions with random bases.
\newblock In \emph{Proceedings of the 46th Annual Allerton Conference}, 2008.

\bibitem[Salomon et~al.(2002)Salomon, King, and Osborne]{SalomonKiOs02}
J.~Salomon, S.~King, and M.~Osborne.
\newblock Framewise phone classification using support vector machines.
\newblock In \emph{Proceedings of the Seventh International Conference on
  Spoken Language Processing}, pages 2645--2648, 2002.

\bibitem[Shalev-Shwartz et~al.(2007)Shalev-Shwartz, Singer, and
  Srebro]{ShalevSiSr07}
S.~Shalev-Shwartz, Y.~Singer, and N.~Srebro.
\newblock Pegasos: {P}rimal {E}stimated sub-{G}r{A}dient {SO}lver for {SVM}.
\newblock In \emph{Proc. ICML 2007}, pages 807--814, 2007.

\bibitem[Shalev-Shwartz et~al.(2010)Shalev-Shwartz, Singer, Srebro, and
  Cotter]{ShalevSiSrCo10}
S.~Shalev-Shwartz, Y.~Singer, N.~Srebro, and A.~Cotter.
\newblock Pegasos: {P}rimal {E}stimated sub-{G}r{A}dient {SO}lver for {SVM}.
\newblock In \emph{Mathematical Programming}, pages 1--34. Springer, October
  2010.

\bibitem[Sonnenburg and Franc(2010)]{SonnenburgFr10}
S.~Sonnenburg and V.~Franc.
\newblock {COFFIN:} a computational framework for linear {SVMs}.
\newblock In \emph{Proc. ICML 2010}, 2010.

\bibitem[Xu et~al.(2004)Xu, Pokharel, Jeong, and Principe]{XuPoJePr06}
J.-W. Xu, P.~P. Pokharel, K.-H. Jeong, and J.~C. Principe.
\newblock An explicit construction of a reproducing {G}aussian kernel {H}ilbert
  space.
\newblock In \emph{Proc. NIPS 2004}, 2004.

\bibitem[Yang et~al.(2006)Yang, Durasiwami, and Davis]{YangDuDa04}
C.~Yang, R.~Durasiwami, and L.~Davis.
\newblock Efficient kernel machine using the improved fast {G}aussian
  transform.
\newblock In \emph{Proc. IEEE International Conference on Acoustic, Speech and
  Signal Processing}, 2006.

\end{thebibliography}



\end{document}